**Title:** Intuitive control of supernumerary robotic limbs through a tactile-encoded neural interface

**Authors:**
Tianyu Jia[1,2]†*, Xingchen Yang[3,1]†*, Ciaran McGeady[1]†, Yifeng Li[1]†, Jinzhi Lin[4], Kit San Ho[2], Feiyu Pan[2], Linhong Ji[2], Chong Li[5]*, Dario Farina[1]*

**Affiliations:**
[1]Department of Bioengineering, Imperial College London, London, UK.
[2]Department of Mechanical Engineering, Tsinghua University, Beijing, China.
[3]School of Automation, Southeast University, Nanjing, China.
[4]School of Biomedical Engineering, Tsinghua University, Beijing, China.
[5]School of Clinical Medicine (BTCH), Tsinghua Medicine, Tsinghua University, Beijing, China.
*Corresponding author. Email: d.farina@imperial.ac.uk (D.F.); t.jia21@imperial.ac.uk (T.J.); chongli@tsinghua.edu.cn (C.L.); xingchen.yang@seu.edu.cn (X.Y.).
†These authors contributed equally to this work.

**Abstract:** Brain-computer interfaces (BCIs) promise to extend human movement capabilities by enabling direct neural control of supernumerary effectors, yet integrating augmented commands with multiple degrees of freedom without disrupting natural movement remains a key challenge. Here, we propose a tactile-encoded BCI that leverages sensory afferents through a novel tactile-evoked P300 paradigm, allowing intuitive and reliable decoding of supernumerary motor intentions even when superimposed with voluntary actions. The interface was evaluated in a multi-day experiment comprising of a single motor recognition task to validate baseline BCI performance and a dual task paradigm to assess the potential influence between the BCI and natural human movement. The brain interface achieved real-time and reliable decoding of four supernumerary degrees of freedom, with significant performance improvements after only three days of training. Importantly, after training, performance did not differ significantly between the single- and dual-BCI task conditions, and natural movement remained unimpaired during concurrent supernumerary control. Lastly, the interface was deployed in a movement augmentation task, demonstrating its ability to command two supernumerary robotic arms for functional assistance during bimanual tasks. These results establish a new neural interface paradigm for movement augmentation through stimulation of sensory afferents, expanding motor degrees of freedom without impairing natural movement.

**One-Sentence Summary:** Tactile-encoded neural interface enables intuitive control of supernumerary limbs without compromising natural human movement

**Main Text:**

# INTRODUCTION

Humans interact with their surroundings with remarkable dexterity and efficiency. Recent advances in robotics and neural interfaces hold the potential to increase these capabilities, enhancing human movement beyond its natural limits. Movement augmentation aims to increase the mechanical degrees of freedom (DoFs) an individual can exert over their surroundings (*1*), allowing movement tasks to be performed more efficiently or enable actions otherwise impossible with natural limbs alone, such as trimanual manipulation with a third arm (*2*). A central challenge, however, lies in achieving practical control of supernumerary effectors (SEs) without compromising natural movement. Here, we present a non-invasive method for SE control that preserves natural movement while augmenting human motor capability.

Current strategies for augmenting DoFs often rely on augmentation by transfer, in which control of SEs is derived from the function of an existing body part, typically one that is task-irrelevant (*1, 3, 4*). For example, a supernumerary thumb has been controlled using a sensor on the toe (*5*). While this approach can increase DoFs for a specific task, it does so at the cost of constraining DoFs elsewhere in the body. By contrast, augmentation by extension aims to increase the body's total number of DoFs by decoding body signals that are independent of overt movement and whose modulation does not interfere with ongoing motor tasks (*1*). Invasive cortical recording techniques have shown promise for cursor control during simultaneous independent natural movement (*6*). While effective, non-invasive approaches such as high-density surface electromyography (EMG) and electroencephalography (EEG) offer a safer, cost-effective, and widely accessible alternative. However, studies investigating redundancy in muscle activity such as those based on null-space muscle activation patterns or high-frequency neural inputs, have shown limited effectiveness and difficulties in the control of supernumerary limbs (*7, 8*).

EEG for control of external devices has been extensively studied within the field of brain-computer interfaces (BCIs) for neurorehabilitation. While rehabilitation primarily aims to restore the lost DoFs following neurological damage (*9*), human movement augmentation pursues the enhancement of unimpaired abilities. Despite these differing objectives, both domains share fundamental principles in interpreting neural activity. Wolpaw *et al.* utilized the modulation of sensorimotor rhythms (SMRs) from EEG to provide one-dimensional (*10*), and later two-dimensional (*11*) cursor control to users with severe motor disabilities. SMRs are oscillatory EEG waveforms found in the alpha/mu frequency range (8-13 Hz) that diminish during motor imagery (MI) and are typically observed over the motor cortical regions (*12*). A BCI can exploit this attenuation to infer user intention by detecting MI of the hand (*13*), foot (*14*), or tongue (*15*). Most of these studies are presented as case reports in which patients used BCIs to operate exoskeletons, functional electrical stimulators (*16*), or wheelchairs. Penaloza & Nishio demonstrated augmentation using an SMR-based BCI, where able-bodied participants activated a robotic arm through MI while performing a bimanual ball-balancing task (*2*). Although interesting, the system provided only a single additional DoF, and the extension to more DoFs presents significant challenges. While MI offers a reliable source of multidimensional neural signals that can be internally modulated, its application in human augmentation is impractical as it requires deciphering supernumerary control intentions that are superimposed on the neural activity associated with natural movement. Another class of signals is aperiodic EEG activity, such as event-related potentials (ERPs), which have been extensively employed in BCI applications. The most studied ERP, the P300, is characterised in EEG as a positive peak of around 5 to 10 microvolts around 200 to 500 ms following onset of a target stimulus. Farwell & Donchin popularised its use with their "P300 Speller", which enabled users to communicate sequences of letters to a computer (*17*). This system provided motor-impaired individuals, such as those with ALS, a novel communication channel (*18*). In this paradigm, patients select characters by attending to a matrix of letters in which rows and columns flash in a random order; the patient focuses on their desired letter, and as the target letter flashes (an infrequent "oddball" event), it elicits a P300 response, which the system detects to infer the intended letter. Typical implementations achieve communication rates of approximately 7.8 characters per minute with accuracies usually above 80% (*19, 20*). The P300 is particularly appealing for human augmentation as it offers a high number of DoFs with little training time and high accuracy; for example, a 6×6 matrix speller enables 36 distinct selectable options. However, its most common implementation, the visual P300, is impractical for movement augmentation as it occupies the user's entire visual attention. A vibrotactile modification to the classic visual paradigm holds promise for augmentation. For instance, Li *et al.* delivered bursts of electrical stimulation to the fingers, analogous to the flashes in the matrix speller (*21*). Users directed their attention to a single finger while P300 responses were decoded using a linear classifier, achieving an accuracy of approximately 80%. Although designed for clinical applications in which the visual system is impaired, we hypothesize that the tactile-evoked P300 could serve as an effective signal for control of artificial supernumerary DoFs for human movement augmentation. This signal is robust, quick to calibrate, and, critically, can be decoded without interfering with natural movements or overloading the visual system.

To test our hypothesis, we investigated whether a tactile-encoded BCI can enable intuitive control of supernumerary robotic arms for movement augmentation. EEG was recorded from ten able-bodied participants while engaged in our tactile-encoded BCI experiment. The BCI leveraged sensory afferents using an oddball paradigm where users selectively focused on vibrotactile target stimuli while ignoring non-targets (Fig. 1). Vibrotactile stimuli were delivered by four tactile vibrators, each corresponding to one BCI command. The P300 response, evoked by the participant's attention to the target stimuli, was detected and used for BCI control. To evaluate the effectiveness of the tactile-encoded BCI for movement augmentation, participants completed three tests: (i) a baseline ball-balancing task to quantify bimanual motor control; (ii) an online BCI test to evaluate the performance of the tactile-encoded BCI for supernumerary command recognition (single-task); and (iii) a dual task combining BCI command recognition with the ball-balancing task to assess the potential interactions between the tactile-encoded BCI and natural movement ability (dual-task). To evaluate potential training effects on BCI performance, the experimental protocol was repeated on three separate days. Lastly, to demonstrate the feasibility of the proposed movement augmentation framework, we designed a functional test where participants controlled two supernumerary robotic arms using the tactile-encoded BCI to complete a four-limb coordination task (two real arms and two supernumerary robotic arms, Fig. S1, Movies S1-S3). Overall, this study investigated (i) whether a tactile-encoded BCI can be utilized for motor intention recognition and DoF augmentation, (ii) whether the performance of the tactile-encoded BCI improves with training, and (iii) the performance of a tactile-encoded BCI used for a functional movement augmentation task.

## RESULTS
### Tactile Vibrator Enables Stable P300 for Motor Intention Recognition

The tactile-evoked P300 over three days of training was evaluated by the amplitude and latency of its positive peak (Fig. 2). For the single-BCI task (Fig. 2A), a one-way repeated-measures analysis of variance (ANOVA) revealed no significant main effect of training time on P300 amplitude ($F_{1.45,13.06} = 0.21$, $p = 0.74$) or latency ($F_{1.62,14.57} = 1.33$, $p = 0.29$). Consistent results were observed in the dual-BCI task (Fig. 2D), with no significant changes in P300 peak amplitude ($F_{1.89,16.99} = 2.27$, $p = 0.14$) or latency ($F_{1.78,16.06} = 0.19$, $p = 0.80$) across three training days.

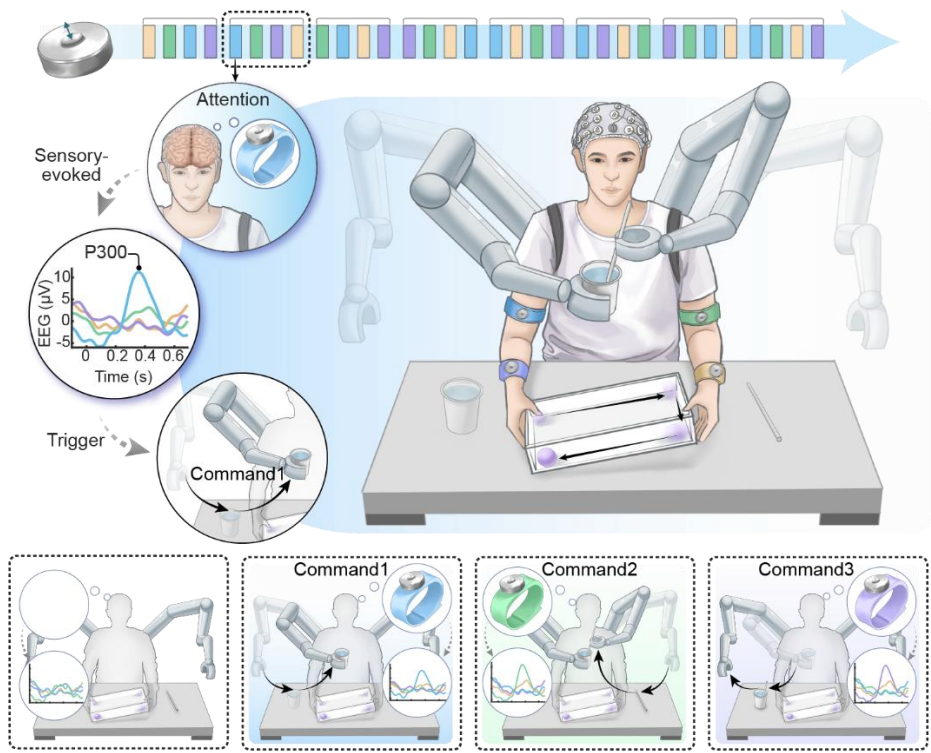

**Fig. 1. Experiment setup for tactile-encoded BCI control of supernumerary robotic arms.** Four tactile vibrators were attached bilaterally to the upper and lower arms using adjustable straps, each delivering 200 ms bursts of vibration with 200 ms interstimulus intervals in a pseudorandom order. Each vibrator corresponded to one of four movement commands for the robotic arm, with symmetric placement to avoid proprioceptive bias. Attended stimuli evoked P300 responses in the EEG, which were decoded online to trigger corresponding robotic commands (Command 1: picks up the cup; Command 2: places the straw in the cup; Command 3: places the cup down). A transparent tray was used for the ball-balancing task in the dual-task condition.

To investigate whether P300 characteristics differed between single- and dual-BCI tasks, we compared the P300 amplitude and latency per day using paired comparisons. There was a significant difference in P300 amplitude between the single- and dual-BCI tasks on all three days [Fig. 2B; day 1 (single: 7.0±2.7 μV, dual: 4.4±1.5 μV, $p < 0.001$), day 2 (single: 6.9±2.7 μV, dual: 3.8±1.8 μV, $p < 0.001$), day 3 (single: 7.2±2.6 μV, dual: 4.7±1.8 μV, $p = 0.001$)]. In contrast, no significant difference was observed in P300 latency on all three days [Fig. 2E; day 1 (single: 349±62 ms, dual: 364±113 ms, $p = 0.77$), day 2 (single: 330±56 ms, dual: 372±116 ms, $p = 0.24$), day 3 (single: 373±51 ms, dual: 388±47 ms, $p = 0.59$)]. Moreover, by analyzing the coefficient of variation (CV) (*22*) for P300 amplitude and latency, we found that there was no significant difference in the CV of P300 amplitude on all three days [Fig. 2C; day 1: $p = 0.87$, day 2: $p = 0.29$, day 3: $p = 0.33$]. Comparatively, there was a significant difference in the CV of P300 latency between single- and dual-BCI tasks on day 1 ($p = 0.006$) and day 2 ($p < 0.001$), whereas not on day 3 ($p = 0.12$) (Fig. 2F). Linear regression analysis of P300 amplitude versus latency CV revealed a significant negative correlation between peak P300 amplitude and latency CV (Fig. 2G and H; single: $R^2 = 0.67$, $p < 0.001$, dual: $R^2 = 0.52$, $p < 0.001$), indicating that lower latency CV reflects higher consistency in P300 responses, thereby enhancing the BCI performance. Therefore, the comparable P300 latency observed between single- and dual-BCI tasks at the last training day (day 3) essentially revealed that both tasks exhibit similar performance after training.

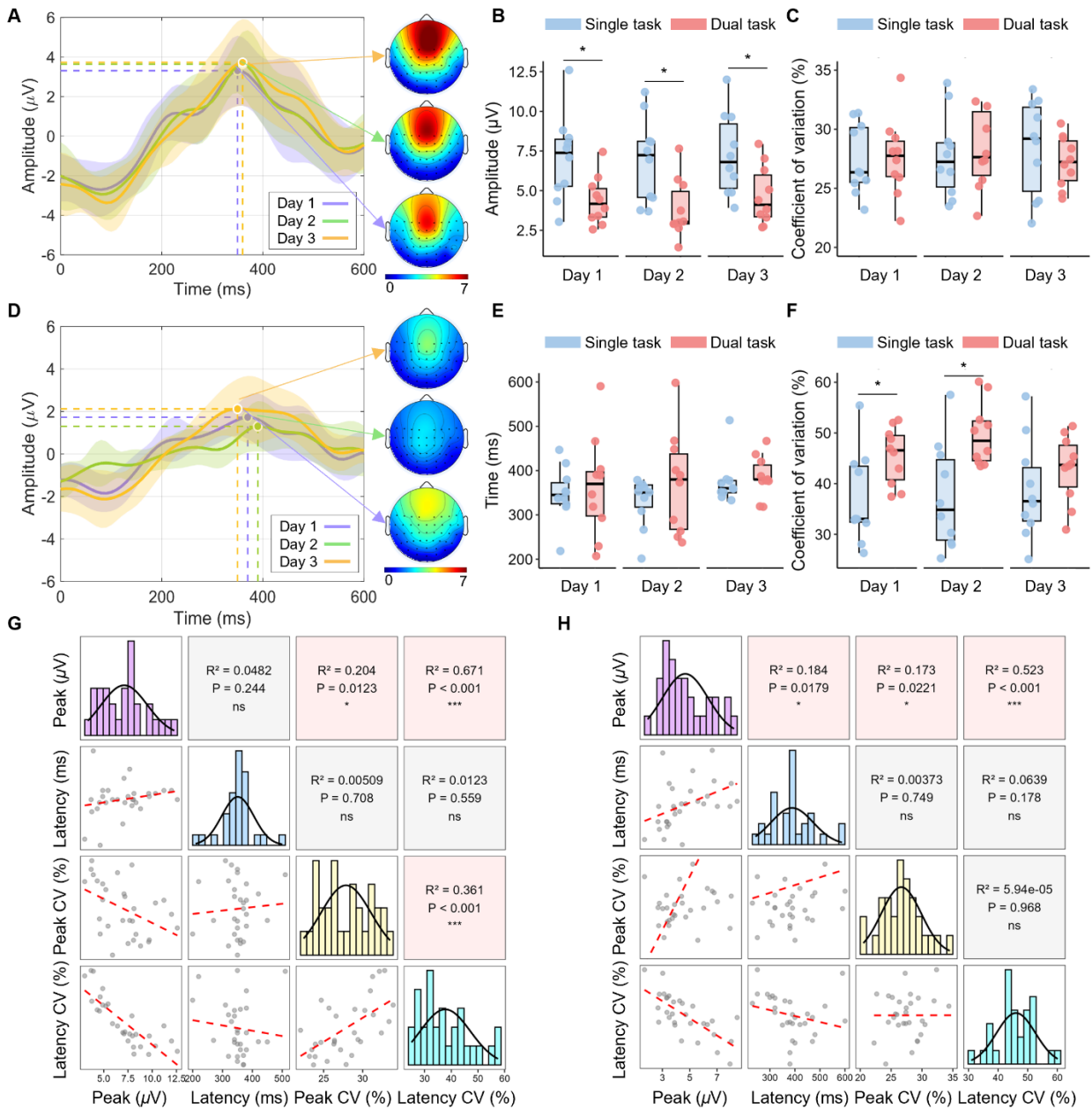

**Fig. 2. Evoked P300 underlying the tactile-encoded BCI tasks over three days of experiment.** Grand-averaged P300 for **(A)** single-task and **(D)** dual-task across days 1-3, with corresponding topographic maps illustrating the P300 peak distribution. The solid lines represent the mean amplitude of the P300, and the shaded boundary represents the standard deviation across participants. Boxplots of P300 **(B)** peak amplitude, **(C)** peak coefficient of variation (CV), **(E)** latency, and **(F)** latency CV across days. Correlation matrices of P300 features for **(G)** single-task and **(H)** dual-task conditions. With an outlier of the peak CV omitted from the figure but included in statistical analysis. (*if $p < 0.05$).

Overall, these results indicate that the additional cognitive demand imposed by the dual task condition was reflected in the reduced amplitude and increased latency CV relative to the single-task condition, suggesting increased neural response variability. Notably, by day 3, the latency variability, measured by CV, converged towards the level observed in the single-task condition, suggesting a potential learning effect with repeated training. The stable P300 evoked by the tactile vibrator would enable accurate recognition of motor intentions.

**Tactile-evoked P300 allows effective movement augmentation and exhibits learning effects**

The online performance of the tactile-encoded BCI was evaluated with two metrics: (i) the percentage of target stimuli that successfully elicited a detectable P300 response (success rate) and (ii) the number of recognition attempts required to successfully detect such a response (response time). Significant improvements in success rates were observed over training for both the single- (Fig. 3A; $F_{1.40, 12.63} = 4.93$, $p = 0.036$) and dual-BCI tasks (Fig. 3B; $F_{1.45, 13.04} = 9.77$, $p = 0.004$). For the dual-BCI task, post hoc analysis with a Bonferroni adjustment revealed that the success rate significantly increased from Day 1 to Day 3 [-16.56 (95% CI, -31.28 to -1.84) %, $p = 0.028$], as well as from Day 2 to Day 3 [-13.44 (95% CI, -24.29 to -2.59) %, $p = 0.016$]. However, for the response time, no significant learning effect was observed for the single- (Fig. 3D; $F_{1.59, 14.27} = 2.74$, $p = 0.11$) or dual-BCI tasks (Fig. 3E; $F_{1.57, 14.15} = 0.03$, $p = 0.95$).

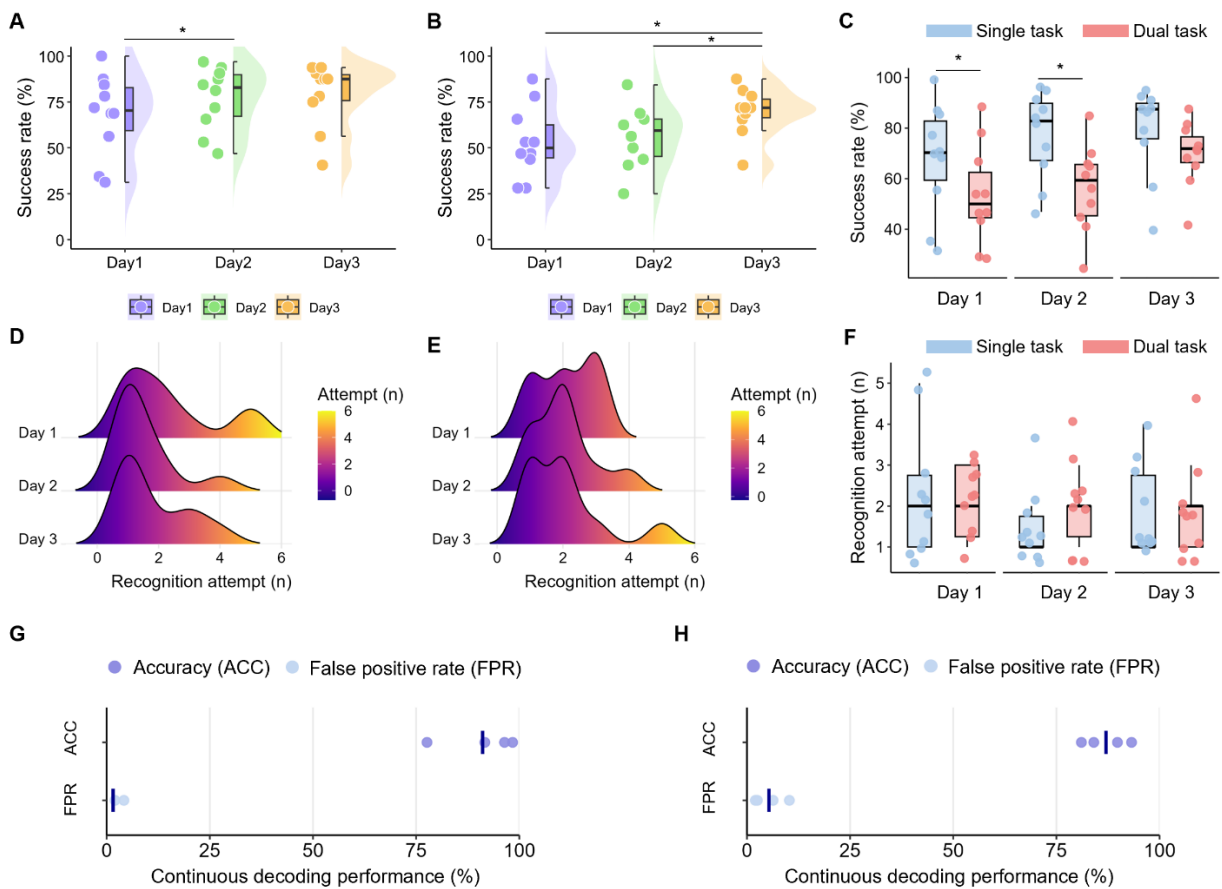

**Fig. 3. Online performance of tactile-encoded BCI over three experimental sessions, comparing single- and dual-task conditions.** Success rates for **(A)** single-task and **(B)** dual-task runs across days 1-3. **(C)** Comparisons of success rate between single- and dual-task conditions on each day. Distribution of the number of recognition attempt before first correct recognition for **(D)** single-task and **(E)** dual-task runs across days. **(F)** Comparisons of recognition attempt between single- and dual-task conditions on each day. Continuous decoding performance metrics for **(G)** single-

task and **(H)** dual-task conditions, expressed as accuracy (ACC), and false positive rate (FPR). Each point represents performance from an individual participant. (* if $p < 0.05$)

To further investigate the difference in BCI performance between single- and dual-BCI tasks, we compared the success rate and recognition attempt per day. Significant differences in success rates were observed on day 1 (single: 68.1±22.1%, dual: 53.1±19.4%, $p = 0.021$) and day 2 (single: 77.2±17.2%, dual: 56.3±16.9%, $p = 0.009$), while not on day 3 (single: 79.1±17.6%, dual: 69.7±12.9%, $p = 0.058$) (Fig. 3C). No significant difference was found in recognition attempt on all three days [Fig. 3F; day1 (single: 2.3±1.6, dual: 2.1±0.9, $p = 0.726$), day 2 (single: 1.5±1.9, dual: 2.0±0.9, $p = 0.273$), day 3 (single: 1.8±1.1, dual: 2.0±1.2, $p = 0.713$)]. Moreover, to validate the sustained decoding ability of the BCI, four participants were recruited for an online continuous decoding test (Fig. 3G and H). All participants could maintain the attention to the target stimulus and sustain accurate recognition of the motor intentions (single: 91.1±9.4%, dual: 87.0±5.6%), with a low false positive rate (single: 1.6±2.2%, dual: 5.3±3.9%). The convergence of success rates between the single- and dual-BCI tasks indicates the learning effect with training, and the sustained detection of enduring and continuous control capabilities of the augmented BCI.

**Tactile-encoded BCI does not interfere with natural human movement**

To assess the impact of the tactile-encoded BCI on natural movement, participants performed a bimanual ball-balancing task that served as a baseline for coordinated motor control. The influence of BCI on natural movement was quantified by ball-balancing performance with and without using the BCI. Task success rate and ball rotation time were quantified as performance metrics. A one-way repeated-measures ANOVA revealed no significant differences across the baseline, dual-task calibration test, and dual-task online test for either success rate [Fig. 4A; day1 ($F_{1.26,11.31} = 2.46$, $p = 0.14$), day2 ($F_{1.87,16.85} = 0.24$, $p = 0.78$), day3 ($F_{1.26,11.35} = 0.62$, $p = 0.48$)] or ball rotation time [Fig. 4B; day1 ($F_{1.69,15.25} = 0.50$, $p = 0.59$), day2 ($F_{1.38,12.41} = 0.45$, $p = 0.58$), day3 ($F_{1.96,17.67} = 0.68$, $p = 0.52$)], indicating that the tactile-encoded BCI did not interfere with natural bimanual coordination.

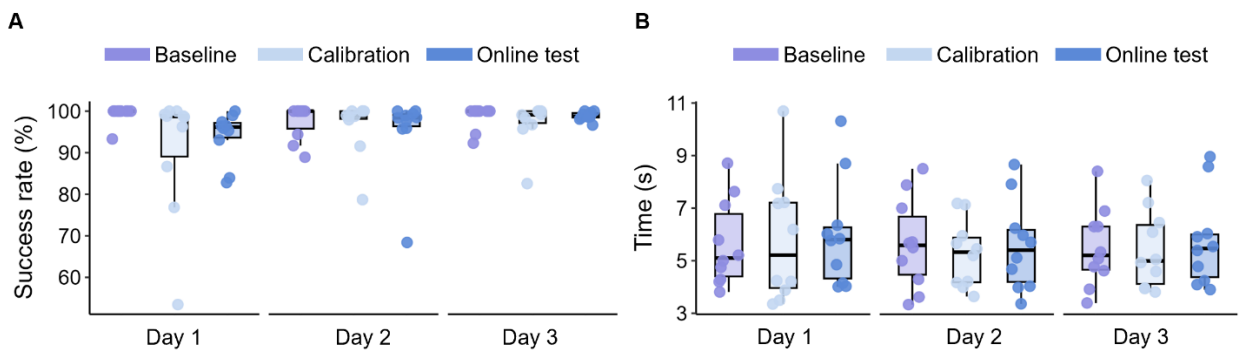

**Fig. 4. Performance of ball-balancing task. (A)** Success rates of ball-balancing task, and **(B)** average time required for each ball rotation during baseline, calibration, and online testing phases across days 1-3. No significant difference was found in the task performance.

**Functional task validation with supernumerary robotic arms**

Three functional tasks were designed to demonstrate the performance of the tactile-encoded BCI for supernumerary robotic arm control (Movies S1-S3). Unlike the virtual online test, which required no transitions between different motion commands, robotic arm control needed sequential and continuous recognition of multiple commands to achieve task objectives. Four tactile vibrators were attached around the upper limbs, each corresponding to one command of the supernumerary robotic arms (Fig. S1). The three movies have been recorded on a representative participant, with performance close to the average performance level of the subject sample.

In the first task, the participant provided unimanual commands to the robotic arms to complete a ball-grasping task (Fig. 5, Movie S1). Starting from their home positions, the robotic arms could be freely selected, either Arm 1 or Arm 2, to perform two possible actions: (i) move to the pickup location and (ii) grasp the ball to return home. This task required both accurate sequential execution and correct arm selection, introducing an additional control challenge. The second task required sequential control of both supernumerary arms to place a straw into a cup while the participant's natural limbs remained engaged in a separate task (Movie S2). The participant first commanded Arm 1 to pick up the cup, then instructed Arm 2 to grasp and drop the straw and finally directed Arm 1 to return the cup to the table. This demonstration highlighted the potential of coordinated bimanual supernumerary control to achieve a shared goal. Finally, in the continuous dual-arm workflow (Movie S3), the arms performed a cyclic sequence in which Arm 1 repeatedly picked up a ball and returned home, while Arm 2 transported the ball to a drop-off point, illustrating reliable repetition of multi-step coordination. Across all demonstrations, the participant performed the tasks smoothly, executing the correct temporal sequence of commands required for successful completion, even while concurrently engaged in natural bimanual movement.

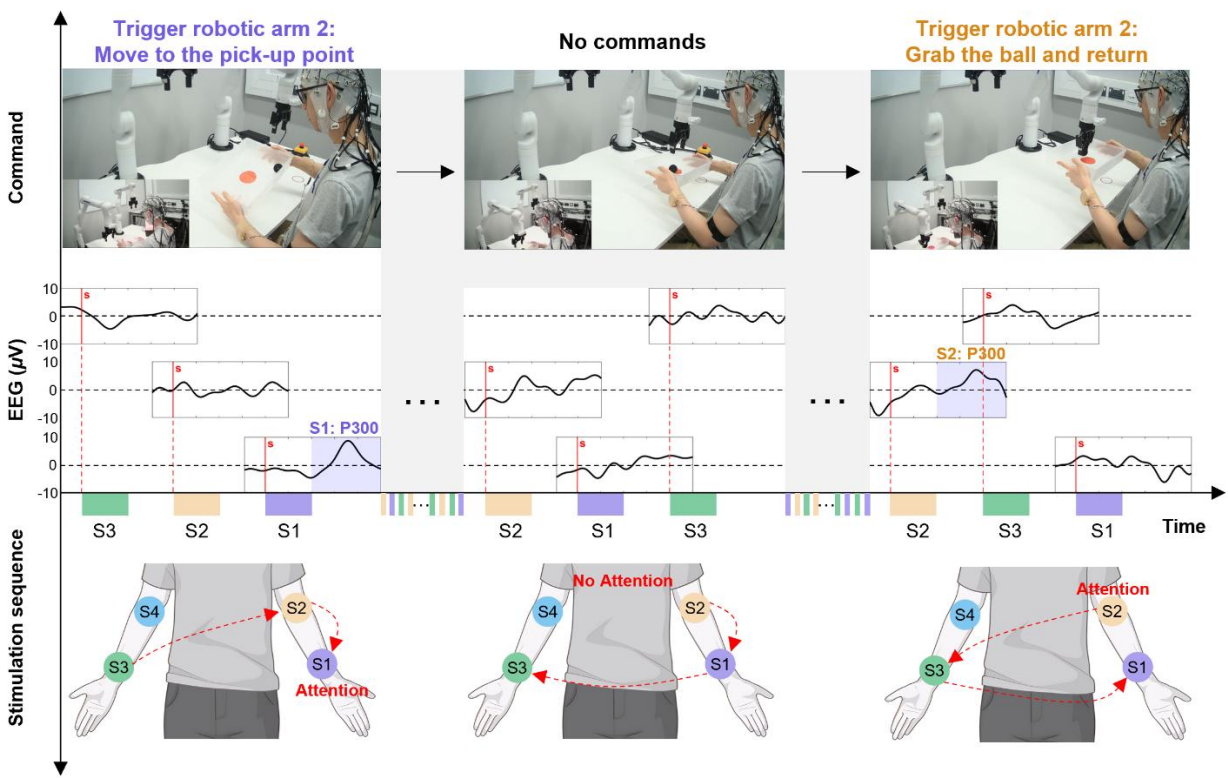

**Fig. 5. Demonstration of online robotic arm control using tactile-encoded BCI.** Representative trial sequence illustrates how tactile-encoded P300 responses were used to control robotic arms. The bottom panel shows the stimulation sequence from vibrators, with the participant's attention directed to the designated target vibrator. The middle panel displays corresponding EEG traces, where clear P300 components (shaded) are elicited by selective attention to the target stimulus (e.g., S1 in the first block, S2 in the third block). The top panel presents the resulting robotic movements: Arm 2 moves to the pick-up point when a P300 is detected for S1, no command is issued when no P300 is identified, and a subsequent P300 for S2 triggers the grasp-and-return action. Together, these sequences illustrate the translation from attended vibrotactile stimuli to robotic control commands in real time.

**DISCUSSION**

In this study, we presented a novel method for supernumerary robotic arm control using a tactile-encoded BCI without compromising natural human movement. Our results first demonstrated the feasibility of evoking P300 through vibrotactile stimulation exploiting the physiological tactile receptors distributed throughout the human body. We then encoded the relationship between tactile perception and motor commands. By combining the evoked P300 with prior tactile encoding, the proposed BCI augmented the controllable DoFs as many as the number of tactile vibrators, without requiring differentiation of overlapping neural activities. In our experiments, we achieved an online task success rate of 69.7% when decoding tasks between four potential targets without compromising the concurrent natural human movement. Additionally, the performance of the BCI significantly improved over three days of training, indicating that practice promoted skill acquisition. This learning effect further promotes the realistic application of our system, as the performance would improve over continuous use.

A central challenge in practical movement augmentation is expanding controllable DoFs without compromising natural movement performance. Previous studies demonstrated that leveraging output-null neural dimensions and kinematic redundancy could provide additional control dimensions for supernumerary limbs. However, achieving stable and intuitive decoupling of augmented and natural motor intentions remains challenging. For instance, foundational studies on superimposing BCI commands over natural movement highlighted this difficulty; while Penaloza & Nishio demonstrated BMI control of a third arm for multitasking (*2*) and Bashford *et al.* achieved concurrent BCI control with overt movements (*6*), this simultaneous control was typically limited to a single degree of freedom. Recent human movement augmentation studies explored control dimensions decoupled from the primary motor task. Dominijanni *et al.* combined gaze and respiration to control a supernumerary robotic arm without influencing the natural limb movements, although the control bandwidth is still limited (*23*); Umezawa *et al.* demonstrated that independent supernumerary limbs could be embodied at the perceptual level, while the controllability remained modest (*24*); Baldi *et al.* systematically reviewed the kinematic redundancy schemes to control supernumerary robotic limbs and concluded that while mapping approaches are feasible, they are often task-specific and difficult to generalize (*25*); A consistent conclusion was achieved by Lee *et al.* from the exploration on null-space muscle activation patterns (*7*). These studies demonstrated that while null-space and kinematic redundancy strategies can, in theory, provide additional control dimensions, their practical use is often constrained by coordination difficulties, the need for substantial user training, and limited scalability. Instead of disentangling motor intentions from superimposed neural activity in a top-down manner, we built a bottom-up approach that encoded neural markers by leveraging sensory afferents through changing attention to vibrotactile stimuli. Our tactile-encoded BCI maps tactile attention to augmented motor output, utilizing the tactile perception of human body to provide supernumerary control dimensions.

Our tactile-encoded BCI enables effective movement augmentation, expanding the control dimensionality by four more DoFs. Although recent non-invasive BCIs have extended online decoding beyond binary control, their performance typically decreases as the number of classes increases. For instance, a tactile P300 system achieved an accuracy of 73% with two vibrotactile targets but dropped to 58% with six targets in a single BCI task (*26*). Likewise, a finger-level EEG decoder using deep learning reached an accuracy of 80.56% for a two-finger motor imagery task but decreased to 60.61% for a three-finger task (*27*). Moreover, only a few multi-class BCI studies have considered the dual-task conditions. Steady-state visually evoked potential (SSVEP)-based BCIs have demonstrated a successful application in a four-class control task during concurrent treadmill walking, with an information transfer rate above 12 bit/min. However, the system relied on gaze-dependent visual stimuli (*28*). Similarly, tactile ERP BCIs have been tested under a multimodal dual-task paradigm, with significant performance degradation while introducing the secondary task (*29*). In comparison, for a four-class decoding task, our system achieved an average accuracy of 74.79% in the single-BCI task and an average accuracy of 59.69% in a dual-BCI task across participants. These online-test results demonstrate that robust four-target online control is feasible even under simultaneous motor demands in a non-visual, tactile paradigm.

The enhancement in BCI performance following training is closely linked to the evolution of the P300 component. Through three days of training, the peak amplitude of P300 under single-BCI tasks was consistently higher than that in dual-BCI tasks. However, no significant difference in P300 latency was found between the single- and dual-BCI tasks across all three days. As we cannot voluntarily modulate the P300 amplitude with training, how did the performance of the dual-BCI task improve with training, and what are the underlying mechanisms that drove this improvement? The decoding performance of the BCI relies on distinguished features, such as the higher P300 peak amplitude and their greater similarity among trials. To quantify this similarity, we introduced CV as a metric. Lower CV values indicate a leptokurtic distribution, reflecting greater consistency in P300 responses across trials, which may enhance decoding performance. As shown in Fig. 2F, by three days of training, no significant difference was observed in the CV of latency between single- and dual-BCI tasks. This indicates that the trial-to-trial variability of the P300 latency tends to converge across both tasks, enabling the model to more reliably capture the key characteristics of the P300 peak at specific time points. Additionally, a significant negative correlation between P300 peak amplitude and latency CV (Fig. 2G and H) further indicated that lower latency variability results in a higher P300 amplitude after trial averaging, which ultimately enhances the performance of the model. We thus hypothesize that through training, participants learned to more effectively allocate attention during dual-BCI tasks, leading to stabilized P300 latency. This stabilization contributed to the improvement of BCI performance.

For human-machine interaction systems, the speed of neural interfaces in decoding motor intentions and outputting commands is of paramount importance. The temporal characteristics of the P300, combined with the need for multiple repeated stimulation to induce a distinct response, introduce decoding delays that can limit responsiveness in quick-response applications (*30*). In our study, after three days of training, most motor intentions could be recognized in the first or second recognition cycle. These results demonstrated that reliable movement augmentation of up to four DoFs could be achieved with minimal latency. However, improving single-trial P300 decoding could further mitigate this latency, but it faces the trade-off between decoding accuracy, computational cost, and real-time feasibility.

De Venuto *et al.* proposed an embedded single-trial P300 decoder that increased the BCI responsiveness, although the validation was limited only to speller data (*31*). Leoni *et al.* introduced an explainable deep learning pipeline that improved single-trial decoding accuracy with increased computational cost (*32*). Du *et al.* reported that fusing multi-subject CNN features boosted the accuracy of single-trial P300 decoding to above 80%, despite requiring large datasets and training overhead (*33*). Kim *et al.* demonstrated a calibration-free "plug-and-play" single-trial BCI using pre-trained spatial filters, showing promising real-time use despite not being tested on tactile paradigms (*34*). Recently, Eidel *et al.* validated tactile P300 with wearable ear-mounted electrodes, but lower channel counts reduced the SNR and efficiency (*35*). In real-time BCI applications, the trade-off between algorithm performance and efficiency is of paramount importance. Therefore, enhancing the performance of single-trial decoding models for ERPs while keeping a high computational efficiency is pivotal to improving the performance of our tactile-encoded BCI.

Neural interfaces are increasingly transitioning from discrete, switch-based control towards continuous control paradigms. For real-time BCI applications, continuous control is desirable compared to a simple action trigger, such as continuous neural control of robotic devices (*36*). Such continuous control relies on the ability of users to voluntarily modulate neural rhythms through real-time neural feedback. For example, for endogenous paradigms such as MI, the modulation of SMRs can be utilized to drive continuous control. In contrast, P300 is an event-related potential elicited by infrequent stimuli. Therefore, its reliance on external stimuli hinders continuous regulation of robotic effectors (*19, 37*), and it was limited within discrete control commands (*17*). As demonstrated by the evolution of P300 features in this study: while P300 exhibited significant differences between single- and dual-BCI tasks due to the variation in attention to the target stimuli (Fig. 2) (*38*), its characteristics cannot be continuously modulated through training. Accordingly, it is difficult to design neural feedback that would allow subjects to voluntarily modulate P300 characteristics for continuous control. To address this limitation and aim at continuous P300 control, we proposed to replace the modulation of P300 characteristics with continuous P300 detection. This approach translates the sustained, uninterrupted detection of P300s into a continuous control signal (e.g., moving a robotic arm between two positions) rather than discrete, event-based commands. By testing the continuous P300 recognition on a subset of four subjects, we showed that the continuous detection had high control accuracy and low false positive rate. This demonstrated that sustained P300 recognition could, to some extent, mitigate the inherent discreteness of the P300 paradigm. Our further work will combine the continuous P300 detection with a ReFit-Kalman filter, enabling continuous and smooth neural control (*39*).

## MATERIALS AND METHODS
### Participants

Ten able-bodied participants (six males, four females, aged between 23 and 37 years) volunteered for this study. All participants were right-handed, possessed normal or corrected-to-normal vision, and had no history of neurological disorders. Two participants had BCI experience, while the others were naïve to BCI experiments. This study was approved by the Imperial College Research Ethics Committees (ICREC reference 18IC4685). Prior to the experiment, participants received an explanation of the experiment procedure and provided signed informed consent. Potential video recording during the experiment was also approved by all participants.

**EEG system and vibrotactile stimulator**

EEG data were recorded at 1000 Hz using a 64-channel EEG cap and an actiCHamp Plus amplifier (Brain Products GmbH, Gilching, Germany) with a customized GUI (Fig. S2) using MATLAB (2024a, The MathWorks Inc., Natick, MA, USA). EEG electrodes were positioned according to the international 10-10 system. A ground electrode was placed on the forehead; electrode impedances were maintained below 5 kΩ. Event markers indicating stimulus onset and vibrator identity were transmitted via a parallel port to ensure precise alignment.

Four C2-HDLF vibrators (Engineering Acoustics Inc., Casselberry, FL, USA) were attached bilaterally to the participants' upper and lower arms using adjustable straps (Fig. S3). Each stimulus consisted of 200 ms bursts of vibration with 200 ms interstimulus intervals, delivered in pseudorandom order. Each vibrator corresponded to one of the four movement commands, respectively. The motors were positioned symmetrically to avoid asymmetric proprioceptive cues.

**Experimental design**

*Baseline ball-balancing phase:* At the beginning of each experimental session, participants were instructed to perform a ball-balancing task to familiarize themselves with the baseline task. Using a lightweight plastic rectangular tray (31.5 × 23.5 cm), they were instructed to roll a lightweight ball sequentially around all four corners at a steady pace. After practice, they were required to perform a 60-second ball-balancing task for baseline task performance assessment. An overhead camera recorded the movement of the ball throughout the session. A rotation was considered successful only if the ball passed through all four corners; incomplete rotations were marked as unsuccessful. The success rate was defined as the percentage of successful rotations relative to the total number of attempts. In addition, the average rotation time was calculated as the total time divided by the sum of successful and unsuccessful rotations, providing an overall measure of performance.

*Exploratory and calibration phase:* Participants were first introduced to the tactile-encoded BCI and the dual-task requirements. To familiarize themselves with the protocol, they completed one single-task and one dual-task exploratory run designed to confirm that all vibration sites were perceived consistently. In each run, one vibrator was designated as the target, and participants silently counted its activations while ignoring non-targets. During dual-task runs, the same procedure was performed while simultaneously performing the ball task. This exploratory phase ensured that participants understood the task demands before the BCI calibration phase. Calibration was conducted separately for the single-task and dual-task conditions. Four vibrators delivered stimuli in a pseudorandom order, with one designated as the attended target. Each run comprised 80 stimuli (20 targets, 60 non-targets), preserving the 1:3 target-to-non-target ratio. Three runs were completed for each of the four target locations. The calibration phase took approximately 15 minutes, and the resulting data were used to train participant-specific decoders for online testing.

*Online evaluation:* Participants completed three experimental sessions on separate days. Each session included both single-task (BCI task only) and dual-task (BCI task combined with bimanual ball-balancing) conditions, with the order counterbalanced across subjects. In single-task runs, participants focused solely on the vibrotactile stimuli, whereas in dual-task runs, they performed the ball-balancing and BCI task simultaneously. For each condition, 32 online tests were conducted, with each of the four vibrators serving as the target eight times in a randomized order. At the beginning of each run, a cue displayed on the screen in front of the participant indicated the target to focus on (Fig. S2). Participants began the ball-balancing task at their own pace following run onset. Feedback was presented on the screen, where the marker corresponding to the current decoding result was illuminated in blue.

Given that the P300 response typically requires trial averaging to achieve a detectable peak, a sliding-window approach was adopted. EEG responses evoked by four consecutive stimuli with the same vibrator were averaged and fed into the decoder. Each stimulation round consisted of four distinct vibration stimuli (one stimulus per vibrator), delivered in a pseudorandom sequence, with a total duration of 1.6 seconds per round. Due to the requirement of trial averaging, the first recognition attempt occurred after the completion of the fourth stimulation round. Thereafter, a decoding decision was made after each subsequent round. Each of the four vibrators corresponded to one BCI command. The classifier output for each vibrator fell into one of two categories: (i) a P300 was detected (triggering the command) or (ii) no P300 was detected (no command output). Online testing continued until one of three outcomes occurred: (i) the target was correctly classified (success), (ii) a non-target was identified (failure), or (iii) all 80 trials were finished (failure). After each run, participants were given a seven-second break before the next run. The number of recognition attempts required until the target P300 was detected was recorded as the response time, and the number of successful online tests out of 32 tests was recorded as the success rate, serving as a metric for evaluating the responsiveness of BCI control. This online evaluation phase modelled the simultaneous control of supernumerary effectors during ongoing motor activity. The experimental protocol was repeated on three separate days to assess potential learning effects on decoding accuracy and P300 waveform modulation. To explore the feasibility of continuous decoding, four out of ten participants completed an additional test in which sustained P300 detections across sliding windows were treated as continuous control inputs. Across 80 trials, participants were instructed to maintain attention on a single target and sustain recognition for as long as possible; the indicator light remained illuminated when consecutive trials were correctly decoded.

**Signal processing and BCI decoder training**

***EEG pre-processing:*** EEG from each run was pre-processed using a standard sequence. Data were epoched from -0.1 to 0.7s relative to stimulus onset. A finite impulse response band-pass filter (1-10 Hz) and a notch filter (48-52 Hz) were applied to remove slow drifts and high-frequency noise. The signals were then re-referenced to the average of electrodes M1 and M2, placed on the mastoid processes, to reduce reference-dependent artefacts. After baseline removal, data were downsampled to 100 Hz. For each trial, the post-stimulus interval of 0.1-0.5s was extracted, as the P300 peaks typically appear around 300 ms (*19*). This window emphasizes the positive deflection while excluding later potentials. Due to substantial contamination from eye blinks and facial muscle artifacts, 18 EEG channels over the frontal regions were excluded from the P300 analysis. Across both single-task and dual-task conditions, the grand-average waveform across all ten subjects revealed that the Cz channel had the maximal P300 amplitude on their first day (Fig. 2A, and D). Consequently, changes in P300 characteristics were evaluated based on channel Cz.

***BCI decoder calibration:*** At the single-trial level, the P300 ERP exhibits a low signal-to-noise ratio (SNR). To improve detectability, sliding-window averaging was applied where EEG responses from four consecutive stimuli at the same vibrator were averaged, thereby reducing random fluctuations. Feature extraction was then performed using the xDAWN algorithm (*40*), which learns spatial filters that maximize the variance of target ERPs relative to overall variance, thereby enhancing the P300 response while suppressing background activity. The resulting spatio-temporal features were standardized and classified using a support vector machine (SVM) with a linear kernel, a method shown to perform reliably in P300 BCIs (*41*). Classification was carried out on trial-averaged data to balance noise reduction and response latency.

*P300 analysis:* We analyzed the P300 event-related potential. For each run, target and non-target trials were averaged respectively within participants and then pooled to generate grand averages across the group. Channel Cz was chosen for analysis because it showed the largest P300 amplitude across all ten participants during the first calibration phase on Day 1, serving as a baseline. Prior studies have reported that tactile P300 amplitudes increase with training (*42*); in this work, we tested whether comparable neurophysiological changes will emerge when the paradigm is applied to movement augmentation.

## Statistics

Statistical analysis was performed using IBM SPSS Statistics and R. For the online BCI performance evaluation, success rate and recognition attempt were analyzed across different days and between single- and dual-task conditions. For the P300 characteristics, P300 peak amplitude and latency were analyzed both across different days and between the single- and dual-task conditions. Furthermore, their CV were compared between the two conditions. A one-way repeated-measures ANOVA with a Greenhouse-Geisser correction was applied with main effects of training on success rate, recognition attempt, and P300 peak and latency. Wilcoxon signed-rank tests or paired-sample t-tests were applied to compare BCI performance and P300 characteristics between single- and dual-task conditions, depending on whether the differences between paired observations met the assumption of normality. Pearson's correlation analysis was performed to examine the relationship of P300 characteristics, including P300 peak, latency and their CV, with two-tailed tests of significance. The statistical significance level was set to 0.05.

## Supplementary Materials
Materials and Methods
Figs. S1 to S6

**Acknowledgments:**
We thank the members of the Neuromechanics and Rehabilitation Technology Laboratory, Department of Bioengineering, Imperial College London, for assistance with data collection, and Y. Liu for contributions to the development of the supernumerary robotic arm setup.

**Funding:** This study was supported by UK Research and Innovation under the UK government's Horizon Europe funding scheme (Grant 10052152, HybridNeuro).

**Author contributions:**
   Conceptualization: TJ, XY, CM, LJ, CL, DF
   Methodology: TJ, XY, CM, YL, DF
   Formal Analysis: TJ, YL, JL
   Investigation: TJ, XY, CM, YL, JL, KSH, FP
   Resources: LJ, CL, DF
   Data curation: TJ, YL
   Writing – original draft: TJ, XY, CM, YL
   Writing – review & editing: LJ, CL, DF
   Visualization: TJ, XY, JL
   Funding acquisition: DF
   Project administration: DF

**Competing interests:** Authors declare that they have no competing interests.

**Data and materials availability:** The main data supporting the results in this study are available within the paper or the Supplementary Materials. Any additional requests for information can be directed to, and will be fulfilled by, the corresponding authors.

# Supplementary Materials for
# Intuitive control of supernumerary robotic limbs through a tactile-encoded neural interface

Tianyu Jia†∗, Xingchen Yang†∗, Ciaran McGeady†, Yifeng Li†, Jinzhi Lin, Kit San Ho,

Feiyu Pan, Linhong Ji, Chong Li∗, Dario Farina∗

∗Corresponding author. Email: d.farina@imperial.ac.uk (D.F.); t.jia21@imperial.ac.uk (T.J.); chongli@tsinghua.edu.cn (C.L.); xingchen.yang@seu.edu.cn (X.Y.)

†These authors contributed equally to this work.

**This PDF file includes:**

Materials and Methods

Figures S1 to S6

## MATERIALS AND METHODS

**Tactile vibrator setup**

Four C2-HDLF vibrators (Engineering Acoustics Inc., Casselberry, FL, USA) were secured bilaterally to the participants' arms using hypoallergenic medical tape (Fig. S3). Vibrators were placed symmetrically above the wrists and biceps to minimize proprioceptive bias and ensure balanced perception across both arms. Each vibrator operated at a fixed frequency of 60 Hz and delivered brief vibration bursts of 200 ms duration with 200 ms interstimulus intervals. Stimuli were presented in a pseudorandom order across the four sites to avoid anticipatory effects and ensure reliable elicitation of event-related potentials. At the beginning of each experimental session, the skin was cleaned with alcohol wipes, and the intended stimulation sites were marked with a skin-safe marker to guarantee consistent placement across days. This procedure minimized day-to-day variability in tactile perception and ensured reproducibility of the evoked responses.

**Online BCI performance evaluation**

In the online BCI test, only the four-target classification paradigm was implemented, with each vibrator corresponding to a distinct command. To provide a broader assessment of classifier performance, one-, two-, and three-target classification outcomes were retrospectively derived from the four-target online decoding output. This approach enabled performance to be interpreted across progressively reduced decision spaces, illustrating how success rates scale as the number of possible targets decreases. Importantly, this analysis did not involve training separate classifiers but was performed solely as a post hoc robustness evaluation, with the four-target condition representing the actual online control setting used in online BCI experiments. Classification accuracy (Fig. S4) and recognition attempt (Fig. S5) were expressed for the one-, two-, three-, and four-target conditions.

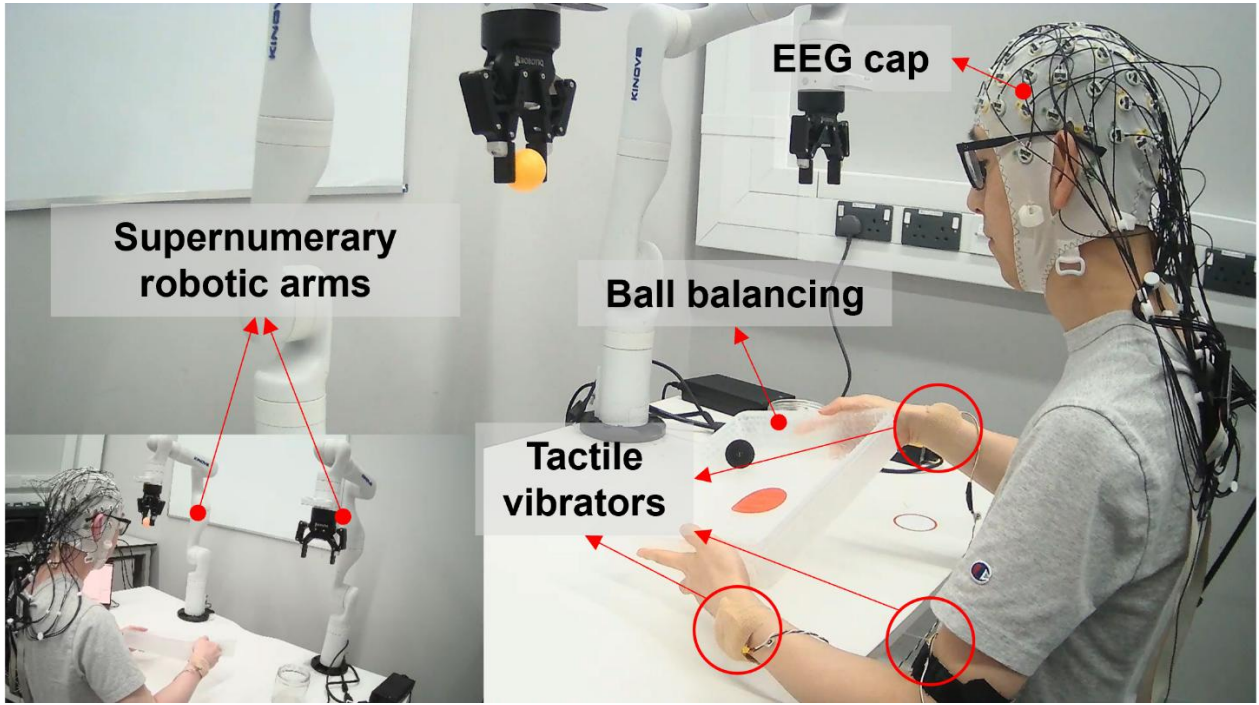

**Figure S1. Experiment setup for the control of supernumerary robotic arms.** Participants wore an EEG cap (actiCAP, Brain Products GmbH, Gilching, Germany) for EEG acquisition while performing a concurrent ball-balancing task. Vibrators were attached to the upper and lower arms to deliver tactile cues, each corresponding to a specific robotic command. Supernumerary robotic arms (Kinova Gen3, Kinova Inc., Boisbriand QC, Canada) executed the decoded commands triggered by the tactile-encoded brain-computer interface in real time, enabling coordinated and augmented movement control.

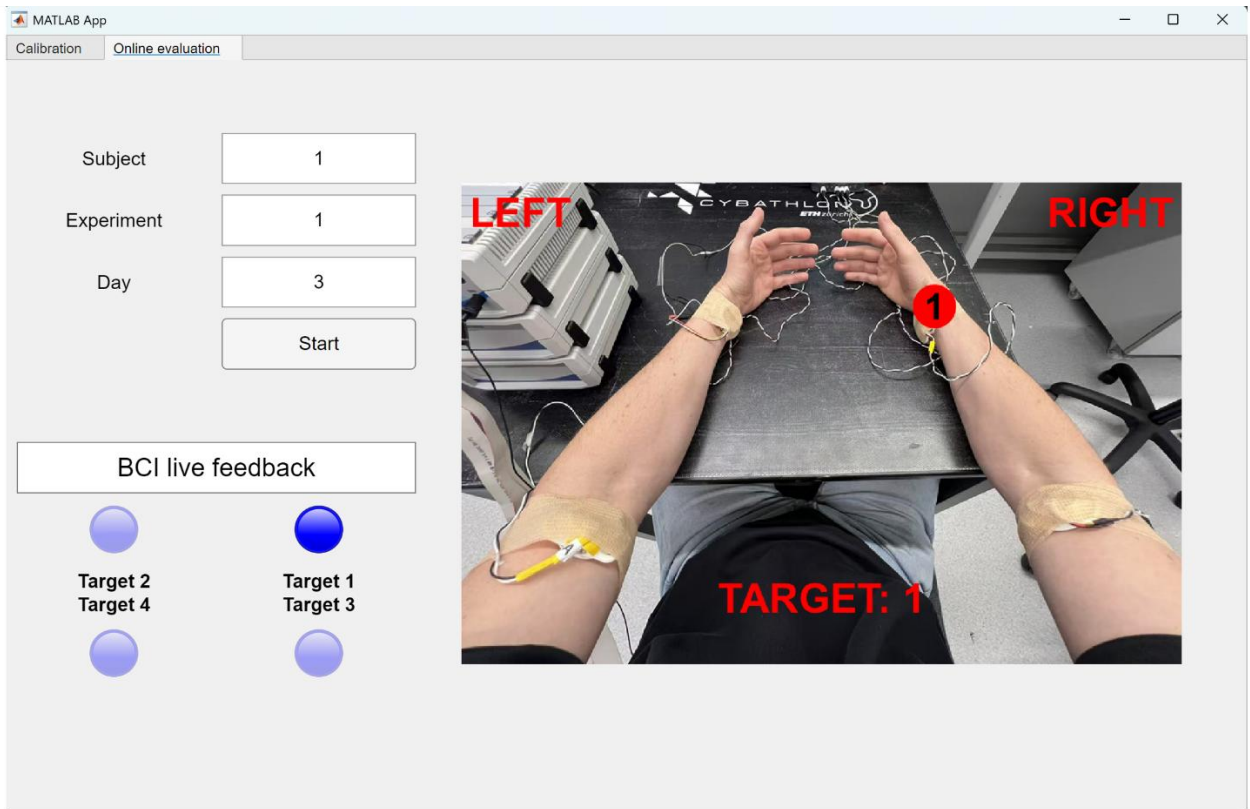

**Figure S2. Customized graphical user interface.** Graphical user interface used for online testing, implemented in MATLAB (2024a, The MathWorks Inc., Natick, MA, USA). The left panel displays subject ID, experiment number, and session day, along with a control button to start the run. The lower left panel provides online BCI feedback by illuminating the marker corresponding to the real-time decoding result. The right panel shows an on-screen cue (TARGET: 1) indicating which vibrator the participant should attend to during the run.

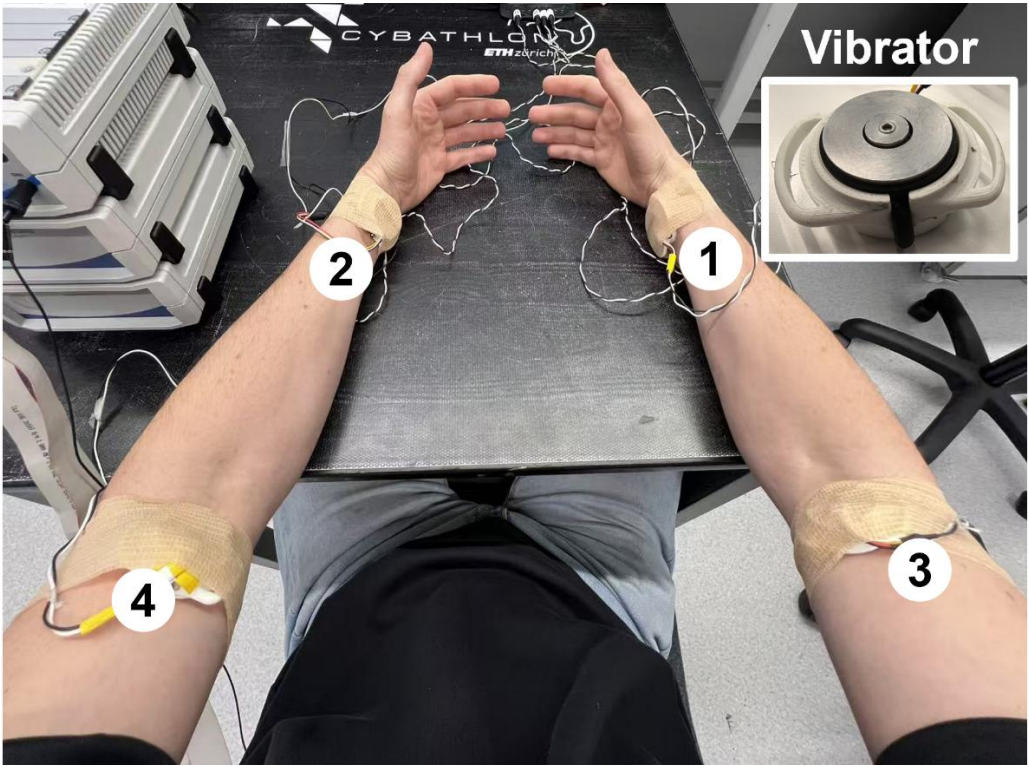

**Figure S3. Experiment setup for vibrotactile stimulation.** Four C2-HDLF vibrators (Engineering Acoustics Inc., Casselberry, FL, USA) were attached bilaterally to the forearms with medical tape. Each operated at 60 Hz and delivered 200 ms bursts with 200 ms interstimulus intervals in a pseudorandom order during the BCI task. The inset shows a close-up of the vibrator.

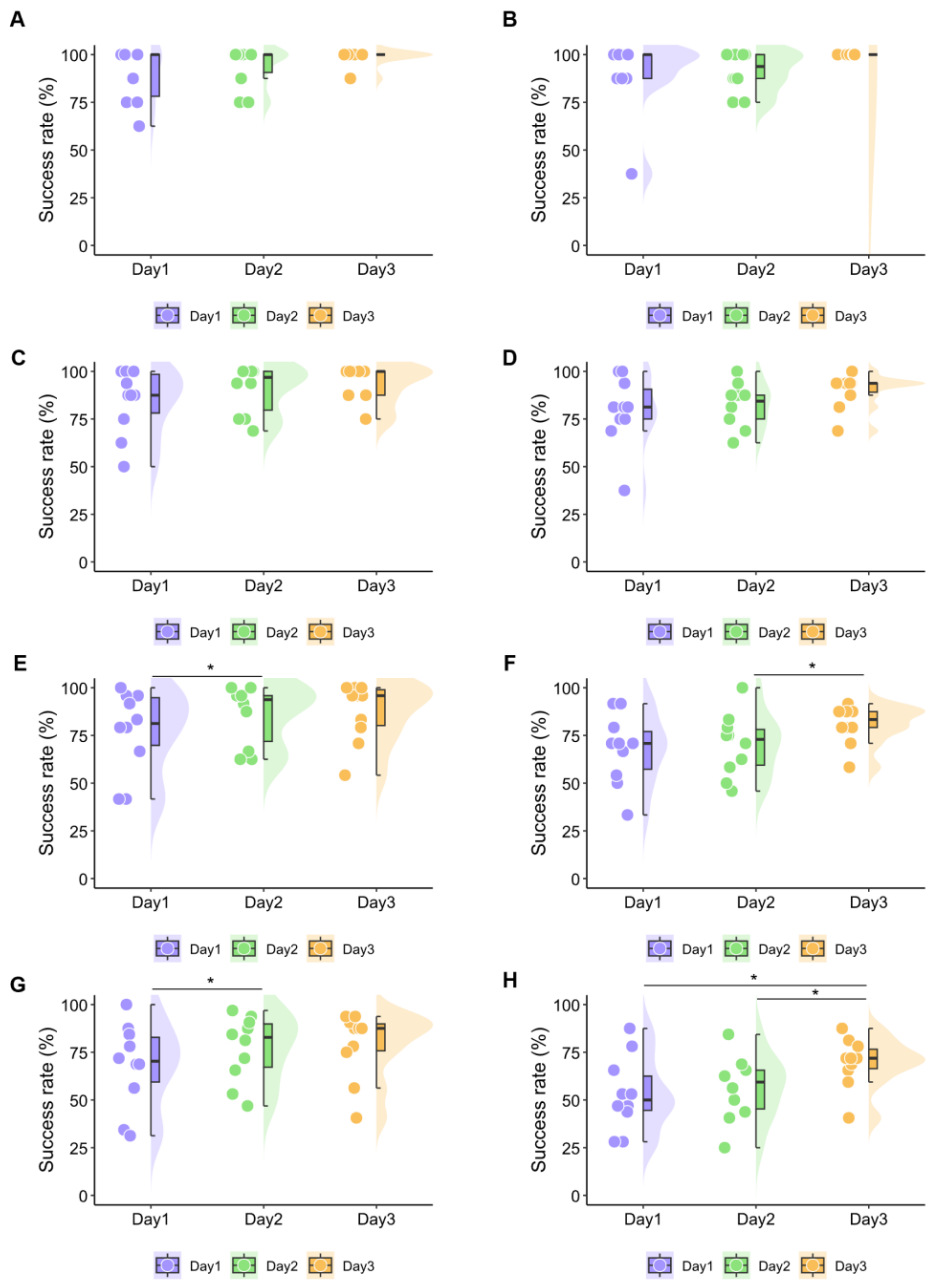

**Figure S4. Success rate distributions of the tactile-encoded brain-computer interface (BCI) across increasing task complexities over three days of training.** Violin plots show participant-level success rates with embedded boxplots. Rows correspond to increasing number of classifiable targets: **(A, B)** one target, **(C, D)** two targets, **(E, F)** three targets, and **(G, H)** four targets. The first column shows single-task (BCI-only) results, and the second column shows dual-task (BCI with ball-balancing) results. Increasing the number of targets reduced performance initially but improvements were seen across days, with significant gains observed in several conditions (* if $p < 0.05$; one-way repeated-measures ANOVA).

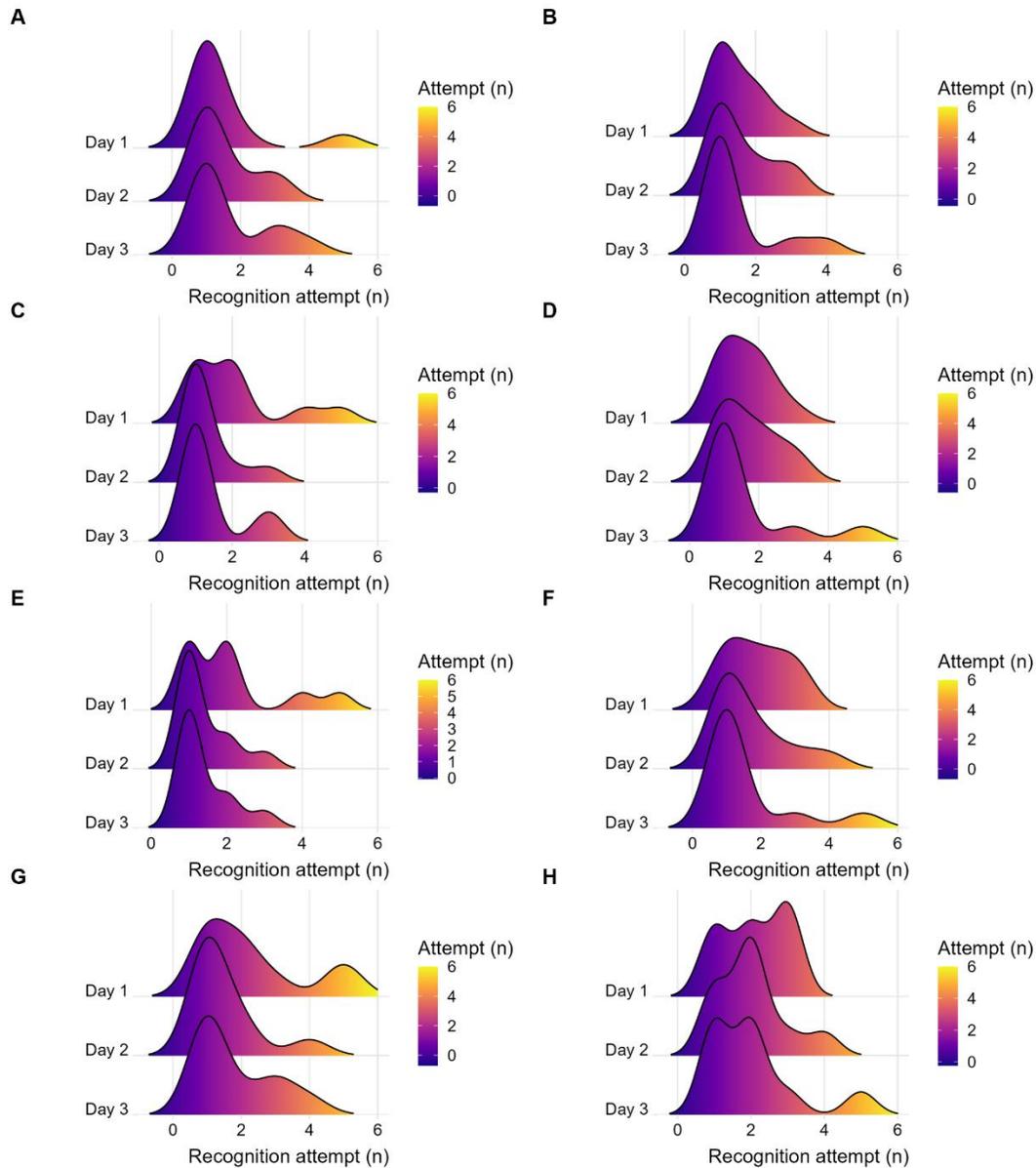

**Figure S5. Recognition attempt distributions of the tactile-encoded brain-computer interface (BCI) across increasing task complexities over three days of training.** Ridge plots showing the distribution of recognition attempts before the first correct classification. Rows correspond to increasing number of classifiable targets: **(A, B)** one target, **(C, D)** two targets, **(E, F)** three targets, and **(G, H)** four targets. The first column shows single-task (BCI-only) results, and the second column shows dual-task (BCI with ball-balancing) results. Data are shown for days 1–3, with colour indicating attempt count (purple = fewer attempts, yellow = more attempts). Increasing the number of targets led to more recognition attempts.

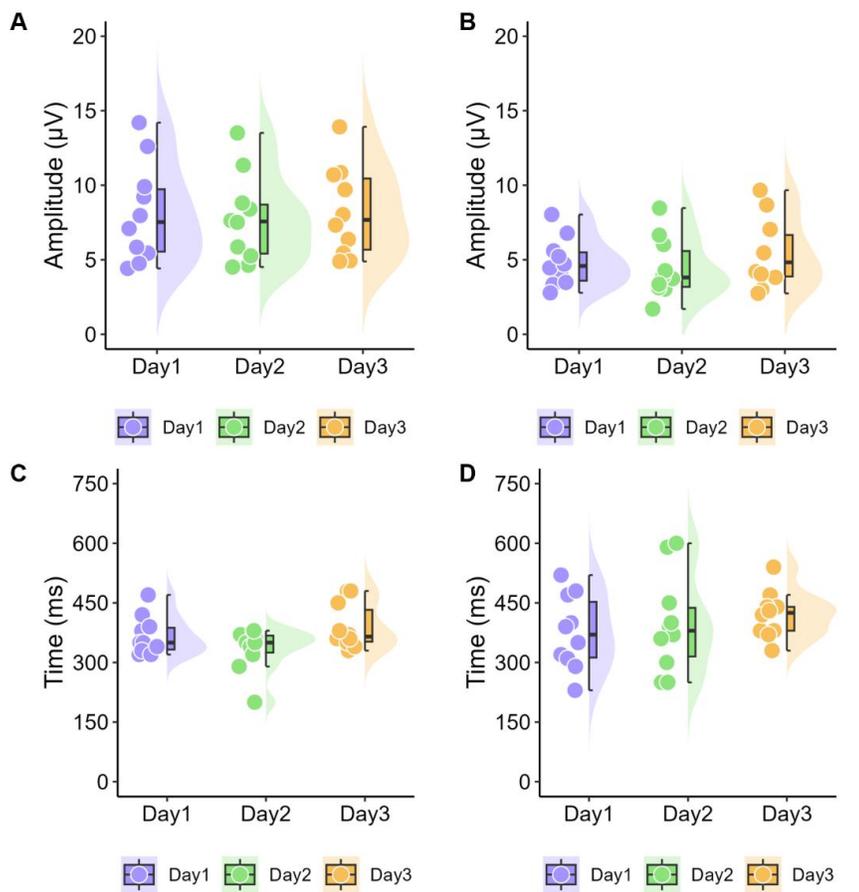

**Figure S6. P300 amplitude and latency over three days of training.** P300 peak amplitude at Cz across days 1-3 under **(A)** single-task and **(B)** dual-task conditions. P300 peak latency across days 1–3 under **(C)** single-task and **(D)** dual-task conditions. Violin plots show participant-level distributions with embedded boxplots. One-way repeated-measures ANOVA showed no significant differences across days for either amplitude or latency.